\pdfoutput=1

\documentclass[11pt]{article}
\usepackage[final]{acl}
\usepackage{times}
\usepackage{array} 
\usepackage{geometry} 
\usepackage{times}
\usepackage{latexsym}
\usepackage{algorithm}  
\usepackage{algpseudocode}  
\usepackage{amsmath}

\usepackage{changepage}
\usepackage{algpseudocode}
\usepackage[T1]{fontenc}
\usepackage{amsmath}
\usepackage{tikz}
\usepackage{ulem}
\usepackage{multirow}
\usepackage{amssymb}
\usepackage{pifont}
\usepackage{booktabs}
\newcommand{\cmark}{\ding{51}}%
\newcommand{\xmark}{\ding{55}}%
\usepackage[T1]{fontenc}

\usepackage[utf8]{inputenc}

\usepackage{microtype}

\usepackage{inconsolata}
\usepackage{marvosym}

\usepackage{graphicx}
\usepackage{enumitem}
\usepackage{pifont}  
\usepackage{tcolorbox}

\newcommand{\corr}{(\Letter)}

\title{ComRAG: Retrieval-Augmented Generation with Dynamic Vector Stores for Real-time Community Question Answering in Industry}

\author{Qinwen Chen$^{\dagger*}$, Wenbiao Tao$^{\dagger}$\thanks{Equal Contribution}, Zhiwei Zhu$^{\dagger}$, Mingfan Xi$^{\ddagger}$, Liangzhong Guo$^{\ddagger}$\\ 
 \textbf{Yuan Wang}$^{\ddagger}$, \textbf{Wei Wang}$^{\dagger}$, \textbf{Yunshi Lan}$^{\dagger}$\corr\\
  $^{\dagger}$School of Data Science and Engineering, East China Normal University \\
  $^{\ddagger}$Alibaba Group \\
  \texttt{\{qwchen, wbtao, 51255903077\}@stu.ecnu.edu.cn}\\ \texttt{\{mingfan.xmf, liangzhong.glz, jingxuan.wy\}@alibaba-inc.com} \\
  \texttt{\{wwang, yslan\}@dase.ecnu.edu.cn}}

\begin{document}
\maketitle
\begin{abstract}
Community Question Answering (CQA) platforms can be deemed as important knowledge bases in community, but effectively leveraging historical interactions and domain knowledge in real-time remains a challenge. Existing methods often underutilize external knowledge, fail to incorporate dynamic historical QA context, or lack memory mechanisms suited for industrial deployment. We propose \textbf{ComRAG}, a retrieval-augmented generation framework for real-time industrial CQA that integrates static knowledge with dynamic historical QA pairs via a centroid-based memory mechanism designed for retrieval, generation, and efficient storage. Evaluated on three industrial CQA datasets, ComRAG consistently outperforms all baselines—achieving up to \textbf{25.9\%} improvement in vector similarity, reducing latency by \textbf{8.7\%–23.3\%}, and lowering chunk growth from \textbf{20.23\%} to \textbf{2.06\%} over iterations.
\end{abstract}

\section{Introduction}
Community Question Answering (CQA) is a collaborative question-and-answer paradigm where users post questions on online platforms (e.g., Stack Overflow\footnote{\url{https://stackoverflow.com/}} and AskUbuntu\footnote{\url{https://askubuntu.com/}}) and community members contribute answers. This paradigm leverages collective intelligence, allowing users to refine answers through voting, commenting, and editing, ultimately enhancing the quality of shared knowledge~\cite{roy2023analysis}.
With the rise of ChatGPT~\cite{achiam2023gpt}, DeepSeek~\cite{deepseek-aiDeepSeekR1IncentivizingReasoning2025}, and other foundation models, Large Language Models (LLMs) have become powerful tools for CQA. However, existing CQA methods focus on static community knowledge, limiting their applicability to real-world scenarios.

We categorize existing CQA methods as follows: (1) Retrieval-based methods: Retrievers or rankers identify the most relevant answers from the community. Question-answer cross-attention networks with knowledge augmentation are utilized for answer selection~\cite{hu2023enhancing}, while structured information is leveraged to enhance ranking~\cite{askari2024answer,ghasemi2024harnessing}. (2) Generation-based methods: LLMs serve
as community experts to answer professional questions. Techniques such as instruction tuning~\cite{yang2023empower}, reinforcement learning~\cite{gorbatovski2024reinforcement} and contrastive learning~\cite{yang2025muppcqa} equip LLMs with domain and community knowledge.

However, these methods have the following weaknesses: 1) They often overlook external domain knowledge, limiting their applicability for domain-specific industrial applications. 2) Real-time CQA presents a continuous stream of questions rather than a static pool, requiring systems to reflect historical interactions. 3) A suitable memory mechanism is needed for real-time industrial CQA, but existing methods overlook this issue.

Domain knowledge and community interaction history play key roles in industrial CQA, shaping the professionalism and relevance of responses, respectively. Thus, the following two key questions require our attention.
\textbf{(Q1) How can we build a CQA system that combines static knowledge with dynamic reflection on the disparate quality of historical answers?}
Domain knowledge serves as an authoritative benchmark, while community QA history links user queries to relevant insights. Combining both enhances LLMs’ ability to generate reliable CQA responses.
\textbf{(Q2) How can a real-time CQA system manage both the rapidly growing volume of historical QA data and the wide variance in response quality?} The evolution of the community leads to varying quality in historical responses due to the open and collaborative nature of CQA. Efficiently identifying, organizing, and leveraging high- and low-quality QA records becomes crucial for maintaining reliable generation.

To tackle these challenges, we propose ComRAG, a retrieval-augmented generation framework that integrates static domain knowledge with dynamic historical QA interactions to enhance real-time CQA in industrial settings. 
In the query phase, ComRAG supports three strategies based on query characteristics: directly reusing answers from high-quality QA pairs, generating responses with reference to high-quality content, and generating responses while explicitly avoiding low-quality ones.
During generation, an adaptive temperature tuning mechanism ensures more confident responses.
In the update phase, the system dynamically manages high- and low-quality CQA vector stores using a centroid-based memory mechanism, optimizing retrieval efficiency for continuous question streams.

In summary, the contributions are as follows.

\begin{itemize}
    \item We propose ComRAG, a novel retrieval-augmented generation framework that jointly integrates static domain knowledge and dynamic community history to address real-time industrial CQA.

    \item We develop a centroid-based memory mechanism for efficient retrieval and an adaptive temperature tuning mechanism for confident generation.

    \item We extensively evaluate our framework on MSQA, ProCQA and PolarDBQA, demonstrating its effectiveness and efficiency for real-time industrial CQA.
\end{itemize}

\section{Related Work}
\subsection{Community Question Answering}
The CQA task centers on improving the relevance and quality of answers. We categorize existing approaches to CQA into two main paradigms: retrieval-based and generation-based.

\textbf{Retrieval-based methods} aim to identify relevant community answers using retrievers or rankers. 
Some enhance answer selection by integrating cross-attention networks with LLM-augmented knowledge~\cite{hu2023enhancing} or incorporate structured metadata into cross-encoder re-ranking~\cite{askari2024answer}. 
Expert finding in CQA is supported by modeling user interactions via topic-based multi-layer graphs~\cite{amendola2024leveraging} while modality-agnostic contrastive pretraining is proposed for aligning code-question pairs~\cite{li2024procqa}. 
Expanding queries and computing translation-based similarity using category-specific dictionaries improve question retrieval~\cite{ghasemi2024harnessing}.

\textbf{Generation-based methods} rely on LLMs acting as community experts to generate answers. 
Prior work explores strategies such as pretraining a small expert model on documentation and CQA data to inject domain knowledge~\cite{yang2023empower}, reinforcement learning from human feedback (RLHF) using community voting signals as rewards~\cite{gorbatovski2024reinforcement}, and aligning LLMs via multi-perspective ranking and contrastive learning to better satisfy diverse user preferences~\cite{yang2025muppcqa}.

Despite strong performance on static benchmarks, existing CQA methods largely overlook the dynamic nature of community content and the inconsistency of historical responses.
\subsection{Retrieval-Augmented Generation}
Retrieval-Augmented Generation (RAG)~\cite{lewis2021retrievalaugmentedgenerationknowledgeintensivenlp} has become a promising framework for enhancing LLMs with external knowledge access. RAG augments the input to generation models by retrieving relevant documents, which helps mitigate hallucinations, knowledge staleness, and limited interpretability~\cite{gao2024retrievalaugmentedgenerationlargelanguage}.
RAG has shown effectiveness across a range of knowledge-intensive tasks, including open-domain QA and summarization~\cite{siriwardhana2022improvingdomainadaptationretrieval}.
Although current RAG implementations rely on static corpora, its retrieval-generation paradigm naturally lends itself to addressing challenges in real-time industrial CQA by enabling fast access to dynamically updated data and supporting the design of customizable retrieval strategies.

\section{Task Definition}

We formally define the task of answering real-time community questions with external knowledge.
Given a collection of documents $D =\{d_i\}_{i=1}^{|D|}$ as the external knowledge, the community questions arrive as a continuous stream.
Suppose that at this moment, we have collected the community history denoted as $H = \{(q_i, a_i)\}_{i=1}^{|H|}$ where the historical questions are associated with the historical responses.
When there is a new question $q$, we can extract the answer $\hat{a}$ either from external knowledge or from the community history to ensure that $\hat{a}$ equals the ground truth answer $a^*$.
Furthermore, we should determine how to organize $(q, \hat{a})$ in the community history $H$ to meet memory constraints and accommodate future follow-up questions.

\section{Our System: ComRAG}
\label{sec:oursystem}

\begin{figure*}[ht]
    \centering
    \includegraphics[scale=0.71]{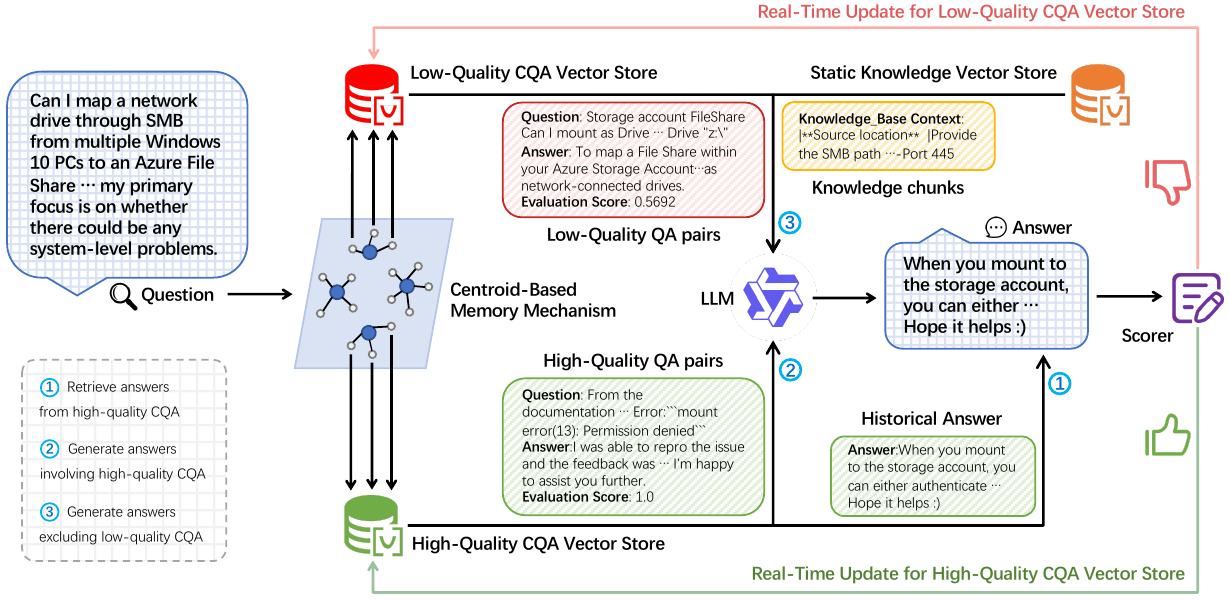}
    \caption{ComRAG architecture for real-time CQA. The system integrates a static knowledge vector store and two dynamic CQA vector stores (high- and low-quality), with the latter managed via a centroid-based memory mechanism. When a question is posed, ComRAG follows one of three paths: \ding{172} retrieving answers from the high-quality CQA vector store, \ding{173} generating answers using high-quality CQA, or \ding{174} generating answers by excluding low-quality CQA and incorporating static knowledge. Real-time updates to either the high- or low-quality CQA vector store ensure efficient memory management and scalable deployment.}
    \label{fig:overview}
\end{figure*}

For real-time CQA in industry, questions can be answered using external knowledge, community history, or a combination of both.
Specifically, the external knowledge is static and filled with domain-specific information.
The community history is dynamic and characterized by accumulated question-answer pairs.
To improve retrieval efficiency and response quality, ComRAG is built upon the RAG framework and contains a \textit{static knowledge vector store} and two \textit{dynamic CQA vector stores}.
The knowledge vector store retrieves relevant domain-specific documents, while the CQA vector stores dynamically maintain and retrieve historical community QA pairs. They work together to either retrieve or generate answers based on high-quality community QA pairs, or alternatively generate answers by avoiding low-quality QA pairs and incorporating external knowledge as additional context.
The overview of ComRAG is shown in Figure~\ref{fig:overview}.

\subsection{Static Knowledge Vector Store}
\label{sec:maincomponents}

Following existing RAG methods~\cite{gao2024retrievalaugmentedgenerationlargelanguage,guo2024lightragsimplefastretrievalaugmented}, we embed the documents in the external knowledge via a static vector database.
Specifically, each document is converted into a vector using an embedding model, allowing us to retrieve relevant documents by computing their similarity to the embedded representation of a given question.
The above procedure can be formulated as follows:
\begin{align*}
    \mathcal{V}_{\text{kb}} &= \{ (d_i, \text{Emb}(d_i)) \mid d_i \in \mathcal{D} \}, \\
    \{\hat{d}_i\}_{i=1}^k  =& \text{arg top-$k$}_{d_i \in \mathcal{D}}{\text{CosSim}(\text{Emb}(q), \text{Emb}(d_i))}.
\end{align*}

Here, $\{\hat{d}_i\}_{i=1}^k$ are the retrieved knowledge documents that serve as evidence of the question. We define $\textit{arg top-k}$ as the function that returns the documents with the top-$k$ similarities.
Then a frozen LLM generates the predicted answers given the instruction and retrieved $\{\hat{d}_i\}_{i=1}^k$ as follows:
\begin{align*}
    \hat{a} = \text{LLM}(q, \{\hat{d}_i\}_{i=1}^k).
\end{align*}

When a question-answer pair is produced, we measure the quality of the answer using a predefined metric, which can be based on either manual or automatic scoring.  Automatic scoring can be implemented using either LLMs or various evaluation metrics, or by combining both.
A score is computed for each QA pair, denoted as $s = \text{Scorer}(q, \hat{a})$.

\subsection{Dynamic CQA Vector Store}
\label{sec:dynamic}

While the static knowledge vector store effectively handles domain-specific questions in industrial settings, it fails to reflect the answers of varying quality in the community history.
Hence, we propose a dynamic Community Question-Answer (CQA) vector store consisting of a \textit{high-quality CQA vector store} and a \textit{low-quality CQA vector store}, which are based on a \textit{centroid-based memory mechanism}.

\paragraph{Centroid-Based Memory Mechanism.}
Since the community history $H$ will continuously increase, we adopt a centroid-based memory mechanism to maintain it within a limited memory size. This mechanism partitions similar historical questions into clusters and only retains the representative questions in each cluster to avoid memory overflow.
Formally, assume we have $m$ clusters $\{C_1, C_2, ..., C_m\}$ in the memory.
Each cluster contains a set of questions belonging to the same topic $C = \{q_i\}_{i=1}^{|C|}$ and its centroid is computed as:
\begin{align*}
   \mathbf{c} = \frac{1}{|C|}\sum_{q_i \in \mathcal{C}} \text{Emb}(q_i).
\end{align*}
Given a new question, we embed it into a vector, then assign it to the most relevant cluster if the similarity exceeds a threshold $\tau$:
\begin{align*}
    C = \text{argmax}_{C \in \{C_1, C_2, \dots, C_m\}} \, \text{CosSim}(\text{Emb}(q), \mathbf{c}).
\end{align*}
The centroid of cluster $C$ is then updated accordingly.

If the similarity is smaller than $\tau$, we believe a question should be derived from a new topic, a new cluster is created in memory with an initial centroid represented as \( \mathbf{c} = \text{Emb}(q) \).

To maintain a fixed memory size, we allow removing questions from a cluster when necessary.
When a newly assigned question is highly similar to an existing one in the cluster (similarity > $\delta$), we compare their answer quality. If the existing question’s answer has a lower evaluation score, we consider the new question to be of higher quality.
Hence, we remove the existing question from the cluster and replace it with the new one.
In this case, we can effectively control the size of each cluster and avoid memory overflow due to the accumulated questions.

Based on this principle, we introduce two vector stores to maintain high-quality and low-quality community history, enabling reflection in follow-up real-time CQA. For high-quality CQA, our system continues to generate similar answers for subsequent questions. For low-quality CQA, our system avoids generating similar answers for follow-up questions.

\paragraph{High-Quality CQA Vector Store.}

We leverage the evaluation score of the answers to decide whether QA pairs are updated into the high-quality or low-quality CQA vector stores.
The high-quality CQA vector store maintains historical QA pairs with scores above $\gamma$, where each answer and its score are stored as metadata.
\begin{align*}
    \mathcal{V}_{\text{high}} = \{ (q, \text{Emb}(q), \hat{a}, s) \mid s \geq \gamma \}.
\end{align*}

To maintain the vector store in a controllable size, we apply the centroid-based memory mechanism to cluster the high-quality question-answer pairs.

\paragraph{Low-Quality CQA Vector Store.}
Similarly, we maintain a low-quality CQA vector store consisting of question-answer pairs with a score lower than $\gamma$ and apply the centroid-based memory mechanism.
\begin{align*}
    \mathcal{V}_{\text{low}} = \{ (q, \text{Emb}(q), \hat{a}, s) \mid s < \gamma \}.
\end{align*}
\subsection{Query and Update}
\label{sec:queryphase}

In the query phase, ComRAG retrieves relevant historical QA pairs and domain knowledge to answer the current question.
The system supports three query strategies:

\begin{enumerate}
    \item[\Large \ding{172}] \textbf{Retrieve answers from high-quality CQA.}
    If the question already exists in the high-quality CQA vector store—that is, the most similar historical question $\tilde{q} = \text{arg max}_{q_i \in \mathcal{V}_{\text{high}}} \text{CosSim}(\text{Emb}(q), \text{Emb}(q_i))$ satisfies $\text{CosSim} \geq \delta$—we directly reuse the corresponding historical answer:
    \begin{align*}
        \hat{a} = \mathcal{V}_{\text{high}}[\tilde{q}].\hat{a}
    \end{align*}
    where $\mathcal{V}_{\text{high}}[\tilde{q}]$ returns the stored tuple $(\tilde{q}, \hat{a}, s)$.

    \item[\Large \ding{173}] \textbf{Generate answers involving high-quality CQA.}
    If the similarity satisfies $\tau \leq \text{CosSim}(\text{Emb}(q), \text{Emb}(\tilde{q})) < \delta$, the retrieved QA pairs still serve as useful references for LLM generation:
    \begin{align*}
        \hat{a} = \text{LLM}(q, \{\mathcal{V}_{\text{high}}[\tilde{q}_i]\}_{i=1}^k)
    \end{align*}
    where each $\mathcal{V}_{\text{high}}[\tilde{q}_i]$ returns the tuple $(\tilde{q}_i, \hat{a}_i, s_i)$, which is used as input evidence. Similar to document retrieval, we select the top-$k$ relevant question-answer pairs as context.

    \item[\Large \ding{174}] \textbf{Generate answers involving low-quality CQA and external knowledge.}
    If no sufficiently similar question is found in the high-quality CQA vector store, we retrieve evidence from both the static knowledge vector store and the low-quality CQA vector store to guide the LLM away from repeating inaccurate historical answers:
    \begin{align*}
        \hat{a} = \text{LLM}(q, \{\hat{d}_i\}_{i=1}^k, \{\mathcal{V}_{\text{low}}[\tilde{q}_j]\}_{j=1}^k)
    \end{align*}
    where each $\mathcal{V}_{\text{low}}[\tilde{q}_j]$ returns $(\tilde{q}_j, \hat{a}_j, s_j)$ for contrastive referencing.
\end{enumerate}

After the predicted answers are generated, we score each answer as described in Section~\ref{sec:maincomponents}, and assign the resulting question–answer pair to either the high-quality or low-quality CQA vector store, as detailed in Section~\ref{sec:dynamic}. Pseudocode for both the query and update phases is provided in Appendix~\ref{appendix:retrieval_algorithm}.

\subsection{Adaptive Temperature Tuning for Generation}
\label{subsec:adaptivetemp}

ComRAG introduces an \textit{adaptive temperature tuning mechanism} to dynamically adjust the LLM's decoding temperature, balancing response diversity and consistency. 
Specifically, for the evidence retrieved from high-quality or low-quality vector stores, we store the answer scores as metadata. If these scores exhibit low variance, this indicates that the historical answers are highly similar; thus we prompt the LLMs with a higher temperature to encourage exploration.
In contrast, when the scores have high variance, we use a lower temperature to ensure consistency with reliable historical answers.

Assume we collect $l$ QA pairs as evidence for prompting, each with an annotated score. After sorting these scores in ascending order \( (s_1, s_2, \dots, s_l) \),  we define the adaptive temperature (with scaling factor $k$) as:
\begin{equation*}
    T(\Delta) = |\exp\bigl(-k \cdot \min_{1 \leq i \leq l-1} (s_{i+1} - s_i))|_{[T_{min}, T_{max}]},
\end{equation*}
where $|\bullet|_{[T_{min}, T_{max}]}$ is a clamp function restricting the temperature to the predefined range $[T_{min}, T_{max}]$.
Then $T(\Delta)$ is set as the argument for the final answer generation.

\begin{figure*}[h]
    \centering
    \begin{minipage}{0.7\textwidth}
        
        \centering
        \small
        \resizebox{\textwidth}{!}{  
        \begin{tabular}{l | c c | c c c c c }
        \toprule
        \multicolumn{3}{c|}{} & \multicolumn{5}{c}{\textbf{MSQA}} \\
        \cmidrule(lr){4-8}
        Methods & Doc & ComQA & Avg Time & BERT-Score & SIM & BLEU & ROUGE-L \\
        \midrule
        Raw LLM         & \xmark & \xmark & \uline{12.70} & 54.70 & 80.58 & 10.46 & 15.07 \\
        BM25        & \xmark & \cmark & 15.07 & 54.98 & 80.41 & 10.44 & 15.03  \\
        DPR         & \xmark & \cmark & 13.91 & 55.01 & 80.67 & 10.65 & 15.09  \\ 
        Vanilla RAG & \cmark & \xmark & 13.86 & 54.43 & 80.73 & 10.09 & 14.82  \\
        RAG+BM25    & \cmark & \cmark & 15.47 & 54.95 & \uline{80.79} & 10.31 & 15.09 \\
        RAG+DPR     & \cmark & \cmark & 14.08 & 55.01 & 80.50 & 10.62 & 15.21 \\
        LLM+EXP     & \cmark & \cmark & 20.23 & \textbf{55.79} & 76.70 & \uline{11.13} & \uline{15.66} \\
        \midrule
        Ours        & \cmark & \cmark & \textbf{11.60}  & \uline{55.47} & \textbf{94.70} & \textbf{11.61} & \textbf{16.66} \\
        \midrule
        \multicolumn{3}{c|}{} & \multicolumn{5}{c}{\textbf{ProCQA}} \\
        \cmidrule(lr){4-8}
        Methods     & Doc & ComQA & Avg Time & BERT-Score & SIM & BLEU & ROUGE-L \\
        \midrule
        Raw LLM         & \xmark & \xmark & \uline{12.77} & 56.16 & 74.88 & 12.17 & 15.49  \\
        BM25        & \xmark & \cmark & 13.99 & 56.21 & 75.68 & 11.41 & 15.81  \\
        DPR         & \xmark & \cmark & 14.11 & 56.08 & \uline{75.73} & 11.20 & 15.56  \\
        Vanilla RAG & \cmark & \xmark & 16.97 & 57.76 & 75.59 & 14.13 & \uline{18.10}  \\
        RAG+BM25    & \cmark & \cmark & 14.11 & 56.20 & 75.30 & 11.22 & 15.69  \\
        RAG+DPR     & \cmark & \cmark & 13.79 & 56.06 & 74.83 & 10.62 & 15.21  \\
        LLM+EXP     & \cmark & \cmark & 22.69 & \uline{58.40} & 67.78 & \uline{14.36} & 16.70 \\
        \midrule
        Ours        & \cmark & \cmark & \textbf{10.42}  & \textbf{58.41} & \textbf{95.31} & \textbf{14.37} & \textbf{18.13}\\
        \midrule
        \multicolumn{3}{c|}{} & \multicolumn{5}{c}{\textbf{PolarDBQA}} \\
        \cmidrule(lr){4-8}
        Methods     & Doc & ComQA & Avg Time & BERT-Score & SIM & BLEU & ROUGE-L \\
        \midrule
        Raw LLM         & \xmark & \xmark & \uline{4.63}  & 60.34 & 93.51 & 1.60 & 9.40\\
        BM25        & \xmark & \cmark & 5.54  & 63.39 & 94.06 & 4.42 & 20.15\\
        DPR         & \xmark & \cmark & 5.45  & 64.01 & \uline{94.08} & 5.72 & 21.52\\
        Vanilla RAG & \cmark & \xmark & 9.67  & 64.78 & 92.27 & 5.21 & 23.45  \\
        RAG+BM25    & \cmark & \cmark & 10.60 & 65.86 & 92.98 & 6.71 & 24.83 \\
        RAG+DPR     & \cmark & \cmark & 22.98 & 66.55 & 93.45 & 6.66 & 28.47\\
        LLM+EXP     & \cmark & \cmark & 8.15  & \uline{67.00} & 90.11 & \textbf{8.04} & \textbf{33.61} \\
        \midrule
        Ours        & \cmark & \cmark & \textbf{3.55}  & \textbf{67.39} & \textbf{96.04} & \uline{7.81} & \uline{30.19} \\
        \bottomrule
        \end{tabular}
        }
        \captionof{table}{Performance comparison of different methods. "Doc" refers to documents retrieved from the static knowledge vector store, and "ComQA" refers to historical QA pairs retrieved from dynamic CQA vector stores. \cmark\ indicates the source is retrieved; \xmark\ indicates it is not.}
        \label{tab:main_results}
    \end{minipage}
    \begin{minipage}{0.28\textwidth}
        \centering
        \small
        \resizebox{\textwidth}{!}{  
        \centering
        \includegraphics[width=\linewidth]{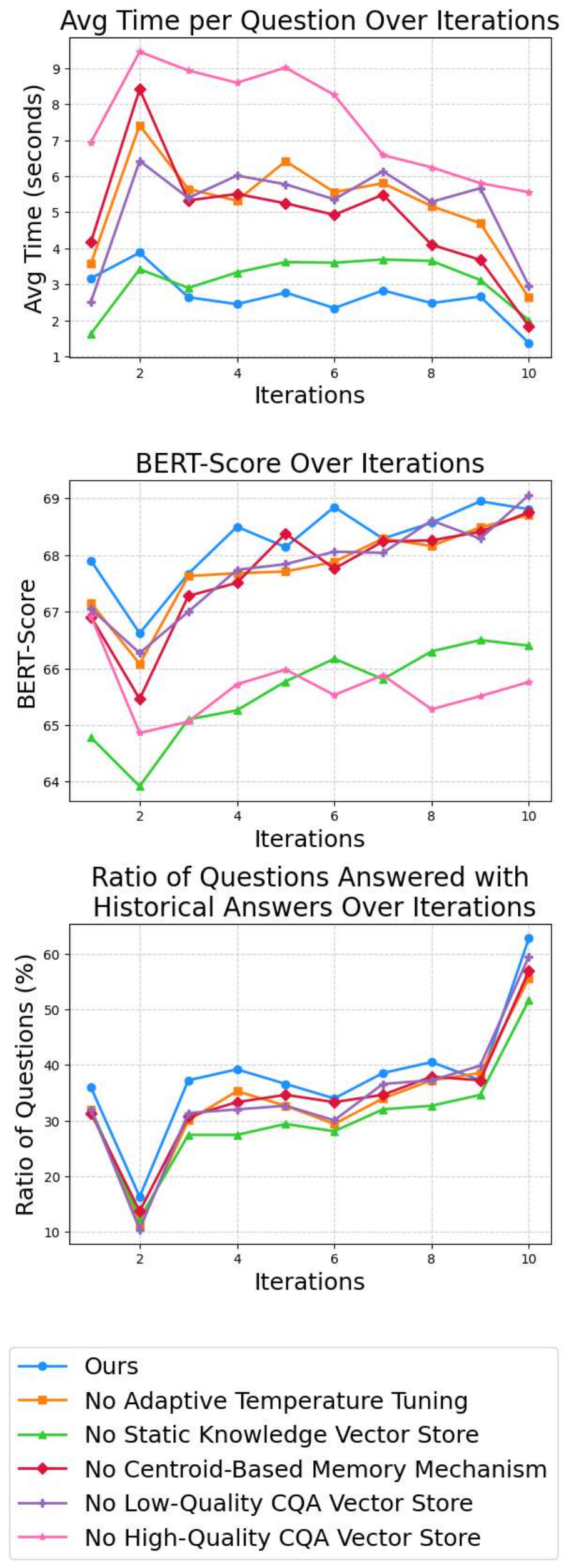}
        }
        \caption{Ablation study on PolarDBQA under a 10-round iterative evaluation setting.}
        \label{fig:ablationstudy}
    \end{minipage}
\end{figure*}

\section{Experimental Setup}

\subsection{Dataset Collection}
\label{subsec:datasets}

We conduct experiments on three community QA datasets: Microsoft QA (MSQA)\cite{yang2023empower}, ProCQA\cite{li2024procqa}, and PolarDBQA. MSQA is a question-answering dataset collected from the Microsoft Q\&A forum. ProCQA consists of structured programming QA pairs extracted from StackOverflow. PolarDBQA is constructed from Alibaba Cloud's official PolarDB documentation, containing question-answer pairs generated by LLM to simulate typical user inquiries in specialized database domains.

For each dataset, there is an associated set of documents as external knowledge. 
MSQA utilizes Azure documentation.
ProCQA provides an official external knowledge corpus for retrieval.
We collect data and construct the PolarDBQA dataset from the Alibaba Cloud platform\footnote{\url{https://docs.polardbpg.com/1733726429035/}, \url{https://help.aliyun.com/zh/polardb/polardb-for-xscale/}}, where PolarDB documentation is utilized as external knowledge.

The question-answer pairs in training sets are initially stored as high-quality CQA vectors.
To simulate real-time CQA where questions arrive sequentially, we paraphrase each question in the test sets into multiple versions using LLMs. We then shuffle all the questions and split them into several iterations for evaluation.
An overview of the datasets is provided in Appendix~\ref{appendix:dataset_stats}.

\subsection{Baselines}

We use \texttt{qwen2.5:14b-instruct-fp16}\cite{bai2023qwentechnicalreport} as the LLM and compare \textbf{ComRAG} with several baselines differing in the external context provided. \textbf{Raw LLM} generates answers without any additional input. \textbf{BM25} and \textbf{DPR} retrieve historical QA pairs using BM25~\cite{10.1561/1500000019} and DPR~\cite{karpukhin2020densepassageretrievalopendomain}, respectively. \textbf{Vanilla RAG} uses only documents retrieved from the static knowledge vector store. \textbf{RAG+BM25} and \textbf{RAG+DPR} extend Vanilla RAG by additionally retrieving historical QA pairs via BM25 or DPR. \textbf{LLM+EXP}~\cite{yang2023empower} follows MSQA’s expert-guided interaction paradigm by aligning knowledge with a domain-specific model and incorporating it into the LLM.

\subsection{Evaluation Metrics}
We evaluate the generated answers using both lexical and semantic metrics. For lexical alignment, we use \textbf{BLEU}~\cite{papineni-etal-2002-bleu} and \textbf{ROUGE-L}~\cite{lin-2004-rouge} to measure n-gram overlap with reference answers. For semantic evaluation, we adopt \textbf{BERT-Score} and report the F1 score. Additionally, we compute the cosine similarity between the embeddings of the generated and reference answers, denoted as the \textbf{SIM} metric. We also report \textbf{Avg Time}, defined as the average processing time per question in seconds.

\subsection{Implementation Details}
\label{sec:comragsettings}
We use the sentence embedding model \texttt{bge-large-en-v1.5-f32}~\cite{bge_embedding} for MSQA and ProCQA, and \texttt{bge-large-zh-v1.5-f32} for PolarDBQA. 
SIM is computed using GPT-2~\cite{radford2019language}, following MSQA. 
For re-ranking, we use \texttt{BAAI/bge-reranker-large}. 
Vector storage and retrieval are managed with ChromaDB v0.6.3. 
All experiments are run on a Linux server with PyTorch 2.6.0 (CUDA 12.4) and Python 3.10.16.
For ComRAG, the core hyperparameters $\tau$, $\delta$, and $\gamma$, introduced in Sections~\ref{sec:dynamic} and~\ref{sec:queryphase}, are set to $(0.75, 0.9, 0.6)$ for MSQA and ProCQA, and $(0.75, 0.8, 0.7)$ for PolarDBQA. 
The scoring function $\text{Scorer}(\cdot)$ used to evaluate answer quality is implemented via BERT-Score~\cite{zhang2020bertscoreevaluatingtextgeneration}, measuring semantic similarity between generated answers and references. 
For the adaptive temperature tuning mechanism, we set $k{=}250$, $T_\text{min}{=}0.7$ and $T_\text{max}{=}1.2$.

\section{Results and Analysis}

\subsection{Main Results}

To ensure a fair comparison under the same initial conditions of real-time QA, we evaluate all baselines and ComRAG on the first iteration of the sequential question-answering setting described in Section~\ref{subsec:datasets}. As shown in Table~\ref{tab:main_results}, ComRAG consistently outperforms all baselines in both answer quality and response efficiency. Compared to the second-best method on each dataset, ComRAG achieves improvements in the SIM metric of \textbf{2.1\%-25.9\%}, and reduces average query latency by \textbf{8.7\%-23.3\%}. These results demonstrate ComRAG’s effectiveness in balancing response quality and latency in real-time CQA.

\subsection{Ablation Study}

We conduct ablation experiments on PolarDBQA under the iterative evaluation setting described in Section~\ref{subsec:datasets}, which simulates real-time CQA by processing questions over multiple rounds. We evaluate the effect of removing each module introduced in Section~\ref{sec:oursystem}. Removing any module increased latency and reduced accuracy, which highlights their necessity as illustrated in Figure~\ref{fig:ablationstudy}. The high-quality CQA vector store had the most significant impact, delaying responses by \textbf{4.9s} and lowering BERT-Score by \textbf{2.6}. Similarly, removing the centroid-based memory mechanism increased delays by \textbf{2.2s} and reduced BERT-Score by \textbf{0.5}, demonstrating its importance in dynamically updating historical QA pairs. Additionally, removing the static knowledge vector store and adaptive temperature tuning mechanism significantly decreased the proportion of directly answerable test questions, indicating that these modules play a crucial role in improving response quality, thereby indirectly enhancing answer reuse efficiency.

\subsection{Real-time QA Evaluation}

\begin{figure}[ht]
    \centering
    \small
    \includegraphics[scale=0.27]{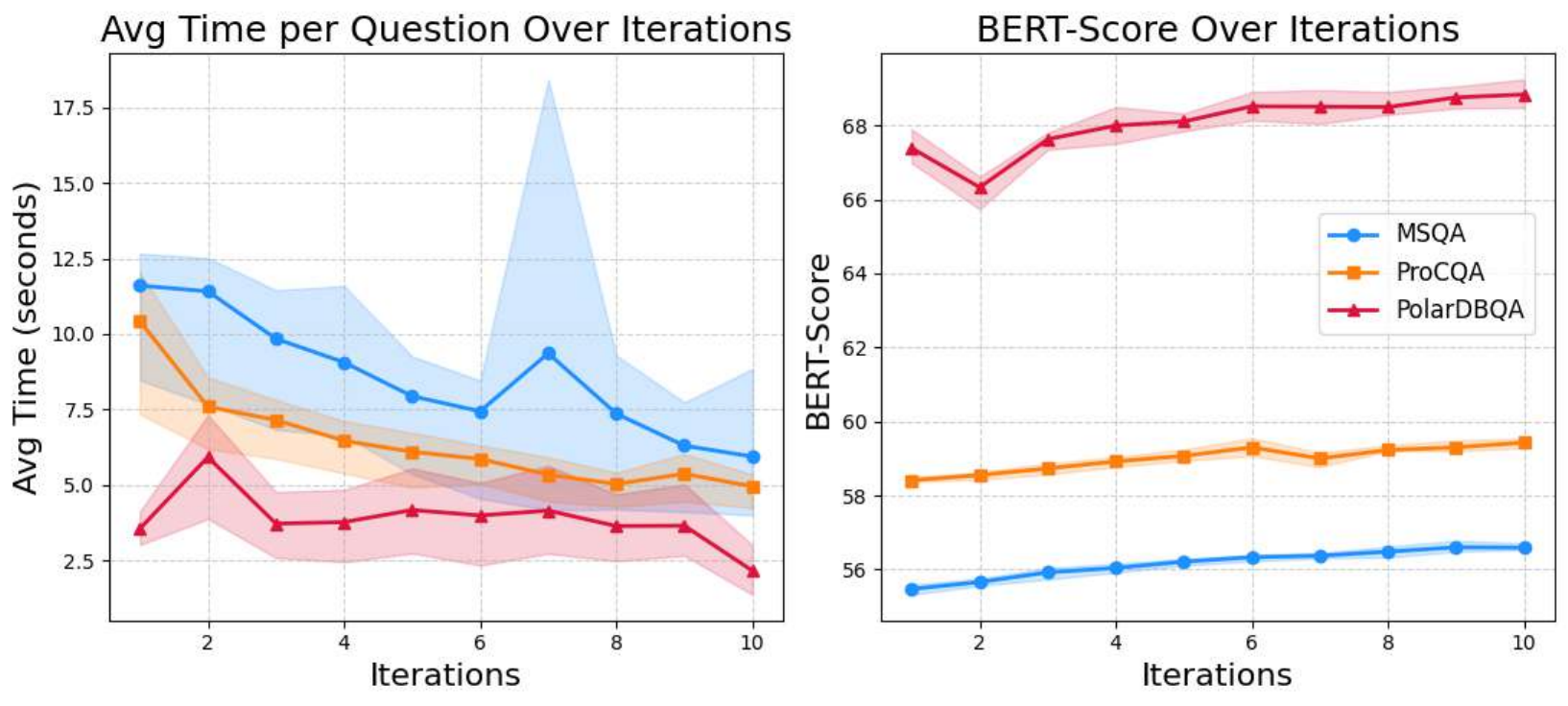} 
    \caption{Avg Time and BERT-Score over iterations. ComRAG improves efficiency and response quality as historical QA interactions accumulate.}
    \label{fig:realtimeqaevaluation}
\end{figure}

We further evaluate ComRAG under the iterative question-answering setting, where questions arrive in sequential batches. As historical QA records accumulate over iterations, ComRAG exhibits consistent improvements in both efficiency and answer quality. As shown in Figure~\ref{fig:realtimeqaevaluation}, query latency drops substantially, with the most notable reduction on ProCQA: average processing time decreases from 10.42s in the first iteration to 4.95s in the final iteration, yielding a \textbf{52.5\%} improvement. Alongside these efficiency gains, response quality also improves, with BERT-Score increasing steadily—most significantly on MSQA, where it rises by \textbf{2.25\%} over time. These results highlight ComRAG’s effectiveness in real-time applications, balancing low-latency generation with progressive quality refinement.

\subsection{Effect of Memory Size}

\begin{figure}[ht]
    \centering
    \small
    \includegraphics[scale=0.28]{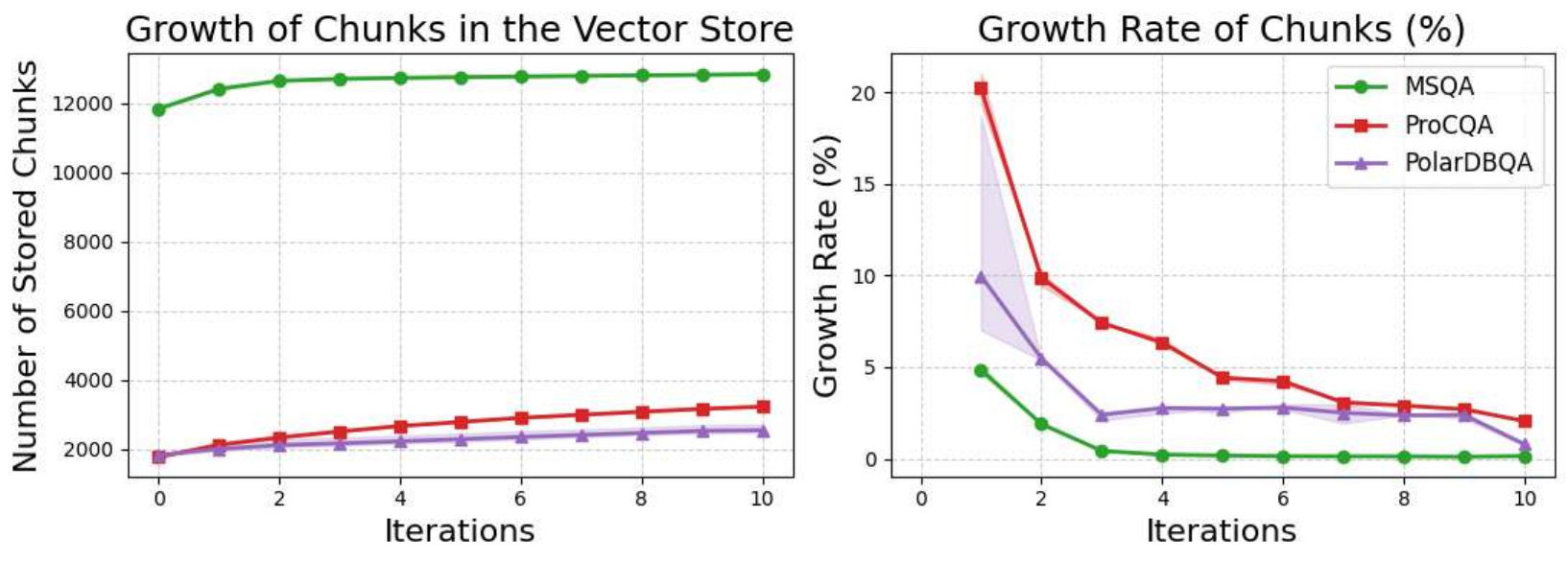}
    \caption{Total stored chunks and growth rate over iterations across all dynamic CQA vector stores. ComRAG efficiently manages memory, preventing excessive storage expansion.}
    \label{fig:memory}
\end{figure}

To evaluate ComRAG’s memory adaptation, we analyze the growth rate of stored chunks across dynamic CQA vector stores over iterations. As shown in Figure~\ref{fig:memory}, the growth rate peaks early and then gradually declines as the system stabilizes. Notably, ProCQA shows the most significant initial expansion, with a \textbf{20.23\%} increase in iteration 1, dropping to just \textbf{2.06\%} by iteration 10. This sharp decline suggests that most necessary knowledge is integrated early, which helps reduce redundant storage in later iterations and stabilizes memory growth over time.  
These results demonstrate that ComRAG effectively manages historical QA storage, preventing uncontrolled expansion while maintaining efficient retrieval. Such controlled memory usage contributes to scalable deployment in real-time industrial CQA systems.




\section{Conclusion}

We present \textbf{ComRAG}, a retrieval-augmented generation framework for real-time industrial CQA. By combining static domain knowledge with dynamic QA history, ComRAG improves response accuracy, latency, and adaptability. It employs a centroid-based memory mechanism to control storage growth and an adaptive temperature tuning mechanism to balance consistency and diversity of generated answers. Experiments on multiple CQA benchmarks demonstrate its practical effectiveness in retrieval and generation for real-world CQA scenarios. 
Furthermore, ComRAG’s modular design supports scalable deployment by enabling replacement of core components such as the LLM backbone, scoring strategy, and retrieval modules, allowing it to accommodate different computational budgets and deployment environments.

\section*{Limitations}
While ComRAG demonstrates strong performance in real-time industrial CQA, several limitations remain. First, the centroid-based memory mechanism relies on fixed similarity thresholds and does not consider topic relevance or usage frequency, which may hinder memory efficiency in dynamic environments. Second, low-quality QA pairs are handled via simple avoidance through prompt design. More advanced filtering or correction mechanisms may enhance reliability. Lastly, the current query and generation paths are rule-based. Incorporating learning-based routing strategies could improve adaptability to diverse question types and knowledge needs.

\section*{Acknowledgments}
The authors would like to thank the anonymous reviewers for their insightful comments. This work is supported by the National Key Research \& Develop Plan (Project No. 2023YFF0725100) and Natural Science Foundation of China (Project No. 62137001 and U23A20298).



\bibliography{custom}


\clearpage  

\appendix

\section{Dataset Overview}
\label{appendix:dataset_stats}

\begin{table}[ht]
    \small
    \renewcommand{\arraystretch}{1.2}
    \centering
    \resizebox{\columnwidth}{!}{
        \begin{tabular}{lccc}
            \toprule
            & \textbf{MSQA} & \textbf{ProCQA} & \textbf{PolarDBQA} \\
            \midrule
            Number of KB Chunks & 557{,}235 & 14{,}478 & 1{,}403 \\
            Train Set Size & 9{,}518 & 3{,}107 & 1{,}395 \\
            Test Set Size & 571 & 346 & 153 \\
            \bottomrule
        \end{tabular}
    }
    \caption{
    Overview of datasets used in our experiments. “Number of KB Chunks” refers to the total number of knowledge base document chunks used as external context. “Train Set Size” denotes the number of QA pairs initially loaded into the high-quality CQA vector store. “Test Set Size” is the total number of test questions evaluated. For ProCQA, we use its Lisp programming language subset.
    }

    \label{tab:dataset_statistics}
\end{table}

\section{Query and Update Algorithms}
\label{appendix:retrieval_algorithm}

Algorithm~\ref{alg:query} outlines the complete query phase in ComRAG.
Given an input question, the system first checks whether any high-quality QA pair in the CQA vector store can be directly reused. If not, it retrieves relevant high-quality QA pairs as references for generation. If no suitable high-quality QA pairs are found, counter-examples are retrieved from the low-quality store, and relevant documents are retrieved from the knowledge base to guide answer generation.

\begin{algorithm}[H]
\caption{Query Phase in ComRAG}
\label{alg:query}
\begin{algorithmic}[1]
    \Require Question $q$, thresholds $\tau$, $\delta$, and $\gamma$  
    , high- and low-quality CQA vector stores  $V_{\text{high}}$ and $V_{\text{low}}$, static knowledge vector store $D$, high-quality centroid vector store $C_{high}$, low-quality centroid vector store $C_{low}$, number of retrieved candidates $k$
    \Ensure Answer \( \hat{a} \)
    
    \State $ \mathbf{q} = \text{Emb}(q) $
    
    \State $\hat{c}_{high} \leftarrow \text{arg top-$k$} \text{CosSim}(\mathbf{q}, C_{high})$ 
    
    \State $\hat{C}_{\text{high}} \leftarrow \text{RetrieveCQA}(V_{\text{high}}, \hat{c}_{high})$
    
    \If {$\text{max}(\hat{C}_{\text{high},i}.sim) \geq \delta$} 
        \State \Return $\hat{a} \leftarrow \hat{C}_{\text{high},i}.answer$  
    \ElsIf {$\tau \leq \hat{C}_{\text{high},i}.sim < \delta$}  
        \State $\hat{a} \leftarrow \text{LLM}(q, \{\hat{C}_{\text{high},i}\})$
    \Else  

        \State $\hat{c}_{low} \leftarrow \text{arg top-$k$} \text{CosSim}(\mathbf{q}, C_{low})$ 
        
        \State $\hat{C}_{\text{low}} \leftarrow \text{RetrieveCQA}(V_{\text{low}}, \hat{c}_{low})$
        
        \State $\{\hat{C}_{\text{low},i}\}_{i=1}^k \leftarrow \text{arg top-k} \text{CosSim}(\mathbf{q}, \hat{C}_{\text{low}})$ 
        
        \State $\hat{D} \leftarrow \text{arg top-k} \text{CosSim}(\mathbf{q}, D)$ 

        \State $\hat{a} = \text{LLM}(q, \{\hat{C}_{\text{low},i}\}_{i=1}^k, \hat{D})$
        
    \EndIf
    \State \Return \( \hat{a} \)
\end{algorithmic}
\end{algorithm}

Algorithm~\ref{alg:update} describes the complete process of the update phase.
After evaluation, ComRAG determines whether the new QA pair should replace an existing entry in the CQA vector store and updates both the vector store and the corresponding centroid. Otherwise, it adds the QA pair to an existing or newly created cluster.

\begin{algorithm}[H]
\caption{Update Phase in ComRAG}
\label{alg:update}
\begin{algorithmic}[1]
    \Require Evaluation result (q,a,s) with the question-answer pair and score, thresholds \( \tau, \delta, \gamma \), high- and low-quality CQA vector stores  $V_{\text{high}}$ and $V_{\text{low}}$, $C$ is the cluster and $c$ is the centroid vector
    \State $ q = \text{Emb}(q) $
    \State $\hat{V} \leftarrow V_{\text{high}}~if~s \geq \gamma~else~V_{\text{low}}$

    \If {$\text{max}(\text{CosSim}(\mathbf{q}, \text{Emb}(\hat{V}_i.q))) \geq \delta$}
        \If {$s > \hat{V}_i.score$}
            \State $\hat{V}.\text{add}((q, \text{Emb}(q), a, s))$
            \State $\hat{V}.\text{delete}(\hat{V}_i)$
            \State $\hat{C} \leftarrow \text{ClusterOf}(\hat{V}_i)$
            \State $\hat{C}.\text{delete}(\hat{V}_i.q)$
            \State $\hat{C}.\text{add}(q)$
            \State $\hat{c} \leftarrow \frac{1}{|\hat{C}|} \sum_{q' \in \hat{C}} \text{Emb}(q')$
        \Else
            \State Discard $(q, a, s)$
        \EndIf

    \Else
        \If {$\text{max}(\text{CosSim}(\mathbf{q}, \text{Emb}(c_i))) \geq \tau$}
            \State $C_{i}.\text{append}(q)$
            \State $c_{i} \leftarrow \frac{1}{|C_{i}|} \sum_{q' \in C_{i}} \text{Emb}(q')$
        \Else
            \State $C_{\text{new}} \leftarrow \{q\}$
            \State $c_{\text{new}} = \text{Emb}(q)$
        \EndIf
    \EndIf
    \State \Return
\end{algorithmic}
\end{algorithm}


\section{Impact of Hyperparameters on ComRAG Performance over Iterations}

\begin{table*}[h]
\centering
\small
\renewcommand{\arraystretch}{1.2}
\begin{tabular}{l p{10cm}}
\toprule
\textbf{Hyperparameter} & \textbf{Role in ComRAG} \\
\midrule
$\tau$ & Used in both the update and query phases: \\
& \quad $\bullet$ In the update phase, it determines whether a new question is similar enough to be assigned to an existing cluster in the centroid-based memory. \\
& \quad $\bullet$ In the query phase, it sets the lower bound for retrieving similar high-quality QA pairs as reference for generation. \\
\midrule
$\delta$ & Used in both the query and update phases to identify near-duplicate questions: \\
& \quad $\bullet$ In the query phase, it decides whether to directly reuse historical answers. \\
& \quad $\bullet$ In the update phase, it determines whether a newly added QA pair should replace a lower-quality one within a cluster. \\
\midrule
$\gamma$ & Used in the update phase to classify QA pairs based on answer quality: \\
& \quad $\bullet$ QA pairs with scores $\geq \gamma$ are stored in the high-quality CQA vector store; others go into the low-quality CQA vector store. \\
\bottomrule
\end{tabular}
\caption{Roles of hyperparameters $\tau$, $\delta$, and $\gamma$ in different phases of ComRAG.}
\label{tab:hyperparam_roles}
\end{table*}
We conduct a series of ablation experiments on the PolarDBQA dataset to analyze the sensitivity of ComRAG to three key hyperparameters: $\tau$, $\delta$, and $\gamma$. Their respective roles in the query and update phases are summarized in Table~\ref{tab:hyperparam_roles}. In each experiment, we vary one hyperparameter while keeping the other two fixed at their default values ($\tau{=}0.75$, $\delta{=}0.8$, $\gamma{=}0.7$). For each setting, we track system performance over 10 iterations using four metrics: Avg Time, BERT-Score, ratio of historical answer reuse, and vector store chunk growth rate.
\label{sec:hyperparameterexperiment}
\begin{figure}[ht]
    \centering
    \includegraphics[width=0.95\linewidth]{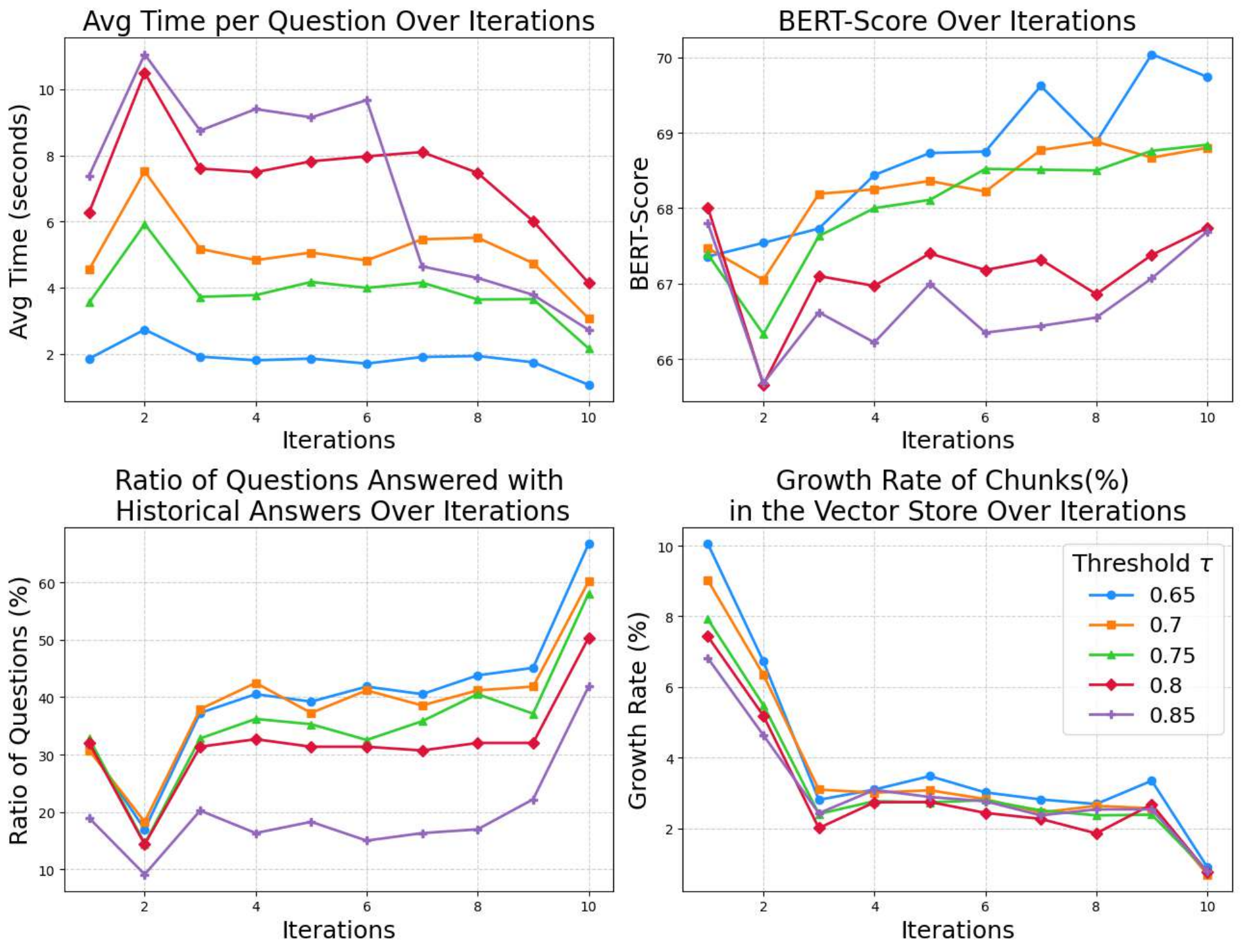}
    \caption{Impact of similarity threshold $\tau$ on ComRAG performance over iterations (with $\delta=0.8$, $\gamma=0.7$ fixed).}
    \label{fig:tau}
\end{figure}
\paragraph{Impact of $\tau$.} 
As shown in Figure~\ref{fig:tau}, lower values of $\tau$ (e.g., 0.65) lead to more aggressive matching with historical QA pairs, resulting in a higher reuse ratio of \textbf{66.67\%} and reduced query latency down to \textbf{1.06s} by the final iteration. This also slows the growth rate of stored chunks, reflecting more efficient memory usage. Conversely, higher thresholds (e.g., 0.85) restrict reuse opportunities, leading to increased latency and memory expansion.

However, overly small $\tau$ values may introduce loosely related historical answers, slightly degrading generation quality as indicated by BERT-Score fluctuations. The default $\tau=0.75$ provides a strong trade-off—ensuring stable semantic quality (e.g., BERT-Score \textbf{68.84}), moderate memory growth, and high efficiency. These findings highlight the role of $\tau$ in balancing reuse, precision, and storage efficiency.

\begin{figure}[ht]
    \centering
    \includegraphics[width=0.95\linewidth]{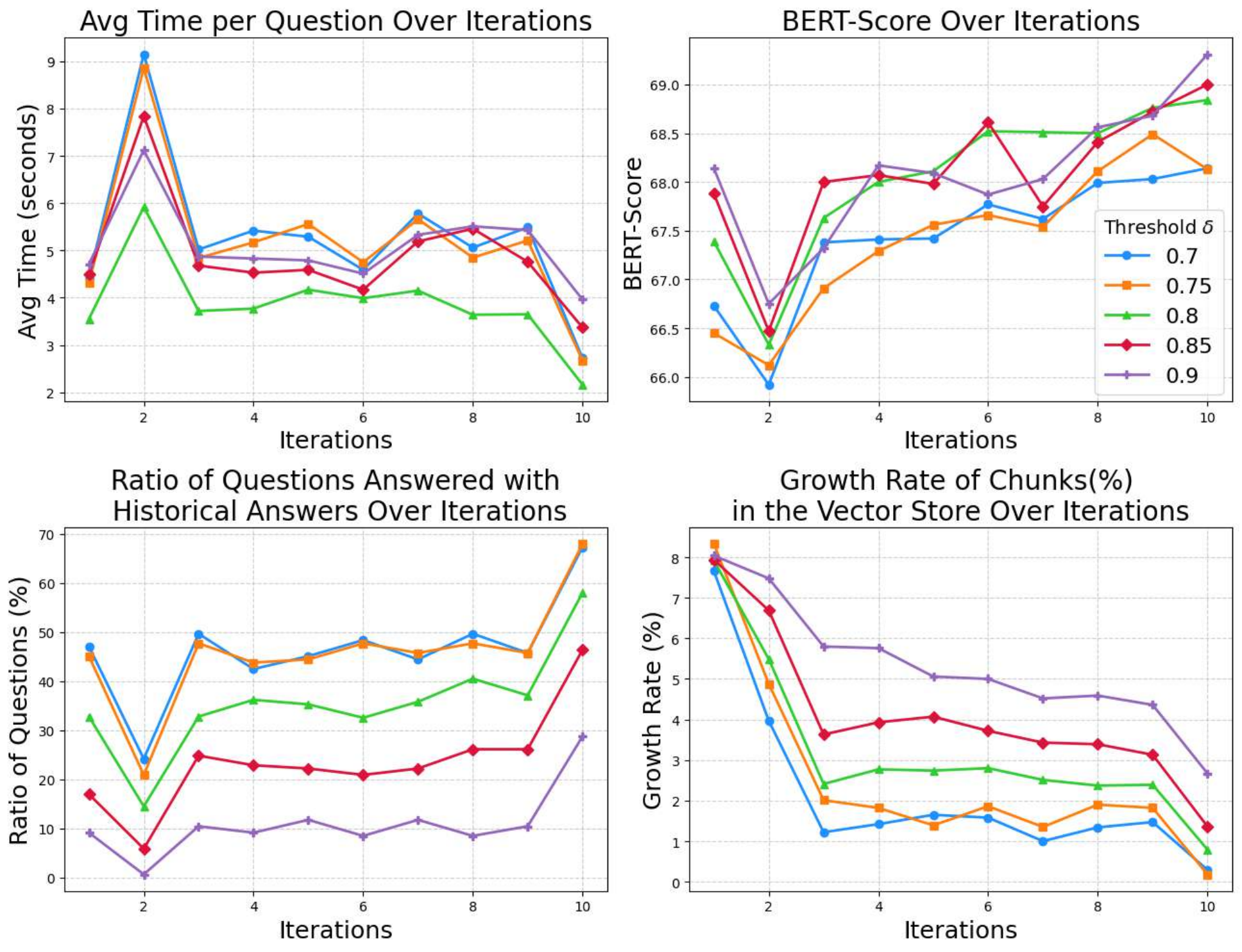}
    \caption{Impact of reuse threshold $\delta$ on ComRAG performance over iterations (with $\tau=0.75$, $\gamma=0.7$ fixed).}
    \label{fig:delta}
\end{figure}
\paragraph{Impact of $\delta$.}
Figure~\ref{fig:delta} shows the impact of $\delta$, which controls the threshold for answer reuse and replacement. Setting $\delta{=}0.8$ yields the best balance across metrics, achieving the highest BERT-Score (68.84), lowest latency (2.16s), and stable chunk growth. A lower $\delta$ (e.g., 0.7) increases reuse ratio (67.32\%) but risks low-quality matches. In contrast, higher values (e.g., 0.9) overly restrict reuse, leading to more generation, higher latency (3.96s), and greater chunk accumulation. This highlights the need for a moderate reuse threshold to ensure both efficiency and quality.
\begin{figure}[ht]
    \centering
    \includegraphics[width=0.95\linewidth]{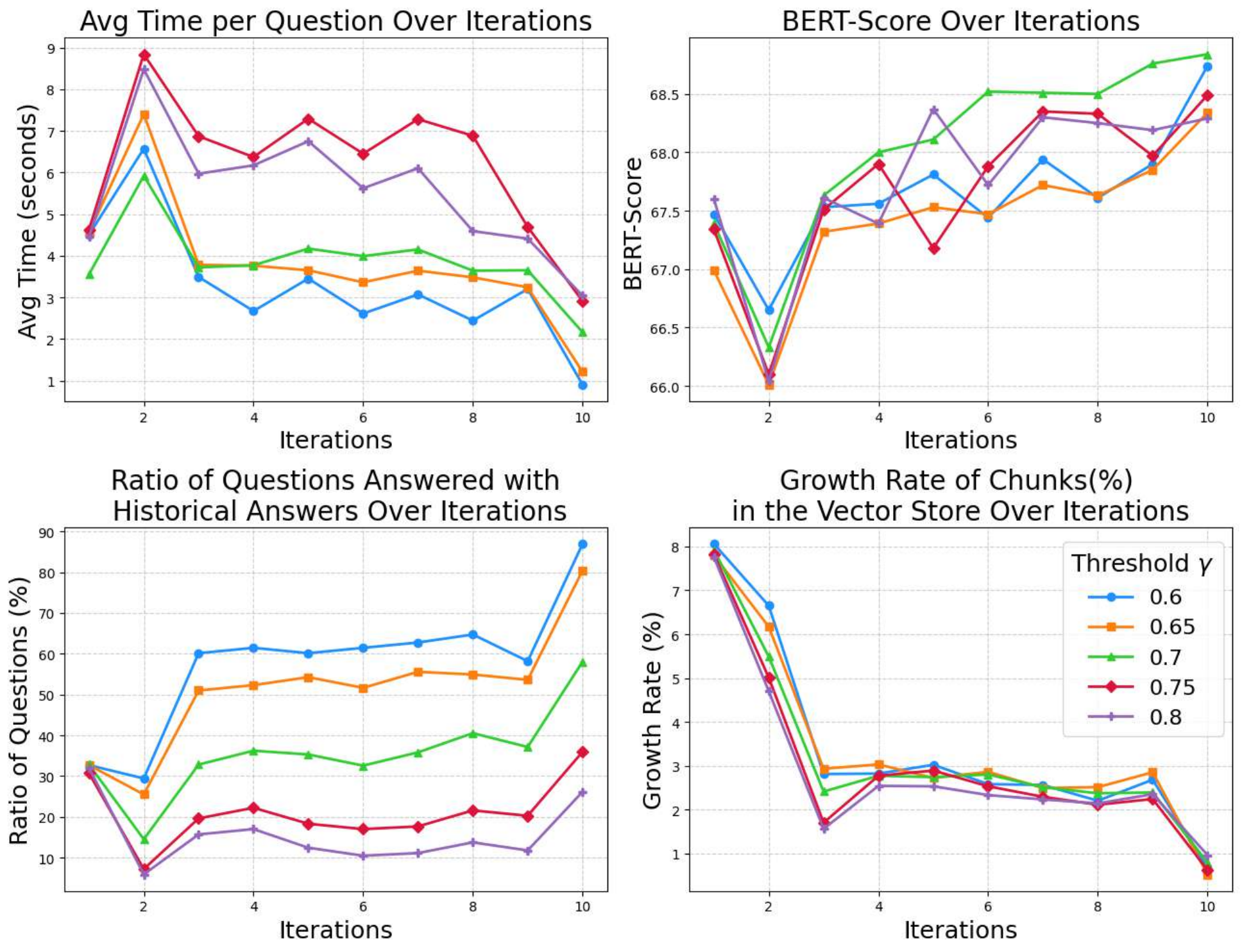}
    \caption{Impact of quality threshold $\gamma$ on ComRAG performance over iterations (with $\tau=0.75$, $\delta=0.8$ fixed).}
    \label{fig:gamma}
\end{figure}

\paragraph{Impact of $\gamma$.}  
As shown in Figure~\ref{fig:gamma}, lower values of $\gamma$ significantly increase the reuse ratio of historical answers—reaching \textbf{86.93\%} at iteration 10 when $\gamma=0.6$, compared to only \textbf{26.14\%} when $\gamma=0.8$. This improvement stems from a relaxed quality threshold for accepting QA pairs into the high-quality CQA vector store, allowing more opportunities for future questions to match and reuse prior answers. As a result, average latency is reduced to as low as \textbf{0.9s}.

However, this comes at the cost of answer quality: lower thresholds admit more low-quality answers, which may be reused directly inappropriately, leading to marginal improvements in BERT-Score. Notably, we also observe an inverse trend in memory growth: higher $\gamma$ values slow the accumulation of stored chunks, as stricter quality criteria make it harder for new QA pairs to enter the high-quality CQA vector store.


\section{Prompts for Answer Generation}
\label{appendix:generation prompt}

We present the prompts for answer generation on MSQA, ProCQA and PolarDBQA in Figure~\ref{fig:prompt msqa}-\ref{fig:prompt polardbqa}.
The prompt instructs the LLM to understand the query, leverage historical QA pairs, utilize domain-specific knowledge sources, handle low-quality historical answers, and output only the answer.

\begin{figure*}
\begin{tcolorbox}[width=1\textwidth, colback=black!5, boxrule=0mm]
    \textbf{\# Role} \\
    You are a proficient expert specializing in answering questions about Microsoft technologies and products, including Azure, Office 365, Windows, and more.\\

    \textbf{\#\#\# System Instructions:} \\
    1. Understand the intent of the question \texttt{previous\_relevant\_qa}: \\
       - Carefully analyze the question to ensure you understand the user's needs.\\
    2. If there is a relevant historical question \texttt{previous\_relevant\_qa}: \\
       If \texttt{previous\_relevant\_qa} is highly similar to the current question, you can directly use the answer from \texttt{previous\_relevant\_qa}.\\
       - If \texttt{previous\_relevant\_qa} is not highly similar to the current question, it can be used as a reference, but the answer should be adjusted to match the current question: \\
          - Based on the feedback score from \texttt{previous\_relevant\_qa}, compare answers with higher and lower scores, and analyze the reasons for improved scores. Avoid repeating mistakes from lower-scored answers to ensure a more accurate answer.\\
    3. If the \texttt{knowledge\_base\_context} exists, the answer should reference it: \\
       - Also, analyze poor Q\&A examples from \texttt{bad\_cqa\_contexts} (if available), comparing answers with higher and lower feedback scores, and analyze the reasons for the improved scores. Avoid repeating errors from low-scored answers, aiming to make the answer as accurate as possible.\\
    4. When there is insufficient context: \\
       - If neither \texttt{knowledge\_base\_context}, \texttt{previous\_relevant\_qa}, nor \texttt{bad\_cqa\_contexts} provide enough information, respond with: “Unable to answer based on available knowledge,” avoiding speculation or providing uncertain information.\\
    5. Provide only the final answer, without including the analysis process.\\

    \textbf{\#\#\# Context} \\
    - knowledge\_base\_context: \texttt{\{knowledge\_base\_context\}} \\
    - previous\_relevant\_qa: \texttt{\{previous\_relevant\_qa\}} \\
    - bad\_cqa\_contexts: \texttt{\{bad\_cqa\_contexts\}} \\

    \textbf{\#\#\# Given Question} \\
    \texttt{\{question\}} \\

    Please return the answer in JSON format, with the structure: \texttt{"answer": "Generated Answer"}
\end{tcolorbox}
\caption{Prompt for answer generation on MSQA}
\label{fig:prompt msqa}
\end{figure*}

\begin{figure*}
\begin{tcolorbox}[width=1\textwidth, colback=black!5, boxrule=0mm]
    \textbf{\# Role} \\
    You are a proficient expert specializing in answering questions about the Lisp programming language.\\

    \textbf{\#\#\# System Instructions:} \\
    1. Understand the intent of the question: \\
       - Carefully analyze the question to ensure you understand the user's needs.\\
    2. If there is a relevant historical question \texttt{previous\_relevant\_qa}: \\
       If \texttt{previous\_relevant\_qa} is highly similar to the current question, you can directly use the answer from \texttt{previous\_relevant\_qa}.\\
       - If \texttt{previous\_relevant\_qa} is not highly similar to the current question, it can be used as a reference, but the answer should be adjusted to match the current question: \\
          - Based on the feedback score from \texttt{previous\_relevant\_qa}, compare answers with higher and lower scores, and analyze the reasons for improved scores. Avoid repeating mistakes from lower-scored answers to ensure a more accurate answer.\\
    3. If the \texttt{knowledge\_base\_context} exists, the answer should reference it: \\
       - Also, analyze poor Q\&A examples from \texttt{bad\_cqa\_contexts} (if available), comparing answers with higher and lower feedback scores, and analyze the reasons for the improved scores. Avoid repeating errors from low-scored answers, aiming to make the answer as accurate as possible.\\
    4. When there is insufficient context: \\
       - If neither \texttt{knowledge\_base\_context}, \texttt{previous\_relevant\_qa}, nor \texttt{bad\_cqa\_contexts} provide enough information, respond with: “Unable to answer based on available knowledge,” avoiding speculation or providing uncertain information.\\
    5. Provide only the final answer, without including the analysis process.\\

    \textbf{\#\#\# Context} \\
    - knowledge\_base\_context: \texttt{\{knowledge\_base\_context\}} \\
    - previous\_relevant\_qa: \texttt{\{previous\_relevant\_qa\}} \\
    - bad\_cqa\_contexts: \texttt{\{bad\_cqa\_contexts\}} \\

    \textbf{\#\#\# Given Question} \\
    \texttt{\{question\}} \\

    Please return the answer in JSON format, with the structure: \texttt{"answer": "Generated Answer"}
\end{tcolorbox}
\caption{Prompt for answer generation on ProCQA}
\label{fig:prompt procqa}
\end{figure*}

\begin{figure*}
\begin{tcolorbox}[width=1\textwidth, colback=black!5, boxrule=0mm]
    \textbf{\# Role} \\
    You are a proficient expert specializing in answering questions about PolarDB. PolarDB for PostgreSQL is a cloud-native database service.\\

    \textbf{\#\#\# System Instructions:} \\
    1. Understand the intent of the question: \\
       - Carefully analyze the question to ensure you understand the user's needs.\\
    2. If there is a relevant historical question \texttt{previous\_relevant\_qa}: \\
       If \texttt{previous\_relevant\_qa} is highly similar to the current question, you can directly use the answer from \texttt{previous\_relevant\_qa}.\\
       - If \texttt{previous\_relevant\_qa} is not highly similar to the current question, it can be used as a reference, but the answer should be adjusted to match the current question: \\
          - Based on the feedback score from \texttt{previous\_relevant\_qa}, compare answers with higher and lower scores, and analyze the reasons for improved scores. Avoid repeating mistakes from lower-scored answers to ensure a more accurate answer.\\
    3. If the \texttt{knowledge\_base\_context} exists, the answer should reference it: \\
       - Also, analyze poor Q\&A examples from \texttt{bad\_cqa\_contexts} (if available), comparing answers with higher and lower feedback scores, and analyze the reasons for the improved scores. Avoid repeating errors from low-scored answers, aiming to make the answer as accurate as possible.\\
    4. When there is insufficient context: \\
       - If neither \texttt{knowledge\_base\_context}, \texttt{previous\_relevant\_qa}, nor \texttt{bad\_cqa\_contexts} provide enough information, respond with: “Unable to answer based on available knowledge,” avoiding speculation or providing uncertain information.\\
    5. Provide only the final answer, without including the analysis process.\\

    \textbf{\#\#\# Context} \\
    - knowledge\_base\_context: \texttt{\{knowledge\_base\_context\}} \\
    - previous\_relevant\_qa: \texttt{\{previous\_relevant\_qa\}} \\
    - bad\_cqa\_contexts: \texttt{\{bad\_cqa\_contexts\}} \\

    \textbf{\#\#\# Given Question} \\
    \texttt{\{question\}} \\

    Please return the answer in JSON format, with the structure: \texttt{"answer": "Generated Answer"}
\end{tcolorbox}
\caption{Prompt for answer generation on PolarDBQA}
\label{fig:prompt polardbqa}
\end{figure*}

\end{document}